%% file: main.tex
\documentclass{article}

\usepackage{arxiv}

\usepackage[utf8]{inputenc} 
\usepackage[T1]{fontenc}    
\usepackage{hyperref}       
\usepackage{url}            
\usepackage{booktabs}       
\usepackage{amsfonts}       
\usepackage{nicefrac}       
\usepackage{microtype}      
\usepackage{lipsum}

\usepackage{booktabs} 
\usepackage{amsmath}
\usepackage{graphicx}
\usepackage{xcolor,colortbl}
\usepackage{enumitem}
\usepackage{etoolbox}
\usepackage{hyperref}
\usepackage{array}
\usepackage{multirow}
\usepackage{makecell}

\begin{document}

\title{Limitations of Pinned AUC for Measuring Unintended Bias}

\author{
Daniel Borkan\\
\texttt{dborkan@google.com}\\
Jigsaw\\
\And 
Lucas Dixon \\
\texttt{ldixon@google.com } \\
Jigsaw\\
\And
John Li \\
\texttt{jetpack@google.com } \\
Jigsaw\\
\And 
Jeffrey Sorensen \\
\texttt{sorenj@google.com }\\
Jigsaw\\
\And
Nithum Thain \\
\texttt{nthain@google.com }\\
Jigsaw\\
\And 
Lucy Vasserman \\
\texttt{lucyvasserman@google.com }\\
Jigsaw\\
}

\maketitle

\input{10_intro}
\input{40_problems_pinned_auc.tex}
\input{70_synthetic_data_example}

\input{90_conclusion}

\bibliography{main}

\bibliographystyle{abbrv}

\end{document}

%% file: 10_intro.tex
\section{Introduction}

This report examines the Pinned AUC metric introduced in \cite{aies_2018} and highlights some of its limitations.
Pinned AUC provides a threshold-agnostic measure of unintended bias in a classification model, inspired by the ROC-AUC metric.
However, as we highlight in this report, there are ways that the metric can obscure different kinds of unintended biases when the underlying class distributions on which bias is being measured are not carefully controlled.
In \cite{aies_2018}, Pinned AUC is applied to a synthetically generated test set where all identity subgroups have identical representation of the classification labels.
This method of controlling the class distributions avoids Pinned AUC's potential to obscure unintended biases.
However, if the test data contains different distributions of classification labels between identities, Pinned AUC's measurement of bias can be skewed, either over or under representing the extent of unintended bias.
In this report, the reasons for Pinned AUC's lack of robustness to variations in the class distributions are demonstrated. We also illustrate how unintended bias identified by Pinned AUC can be decomposed into the metrics presented in \cite{fates-2019}.
To avoid requiring careful class balancing, which is hard to do on real data, instead of using Pinned AUC, the threshold agnostic metrics presented in \cite{fates-2019} can be used; these are robust to variations in the class distributions and provide a more nuanced view of unintended bias. 

%% file: 40_problems_pinned_auc.tex
\section{Pinned AUC Limitations}

\subsection{Pinned AUC Summary}

Pinned AUC provides a threshold-agnostic measure of the extent of unintended bias by measuring model quality for a specific subgroup \emph{within the context of a general background distribution} containing other subgroups.

Pinned AUC for a specific dataset, $D$, and subgroup, $g$, is calculated by creating a secondary, "pinned" dataset made up of 50\% examples within subgroup $g$ and 50\% examples randomly sampled
\[
pD_g = s(D_g) + s(D), \quad |s(D_g)| = |s(D)|
\]
and then calculating the ROC-AUC
\[
p\textup{AUC}_g = \textup{AUC}(\textup{pD}_g)
\]
of that "pinned" dataset.

\subsection{Restriction on Pinned AUC} \label{subsec:pinned_auc}

To understand where Pinned AUC can be effective and where it fails, consider the examples in Table~\ref{table:dist}.
Column A of the table shows hypothetical score distributions produced by a model with unintended bias.
These model scores represent the probability that an online comment will be considered "toxic", where a toxic comment is defined as a comment that is \textit{rude, disrespectful, or unreasonable that would make someone want to leave a conversation}.
The top chart shows scores for the hypothetical background data and the lower chart shows distributions for one hypothetical identity subgroup.
The left (red) distribution represents non-toxic examples and the right (blue) distribution shows toxic examples. 
In column A, the subgroup scores are higher than those of the background for both toxic and non-toxic examples.
In fact, they are so much higher, that non-toxic examples within the subgroup have similar scores to toxic examples in the background, making it impossible to draw a single threshold. 
This score shift is a clear example of unintended bias, and that bias is identified by the lowered Pinned AUC score of 0.94. 

Columns B and C of Table~\ref{table:dist} show model scores that are identical to those in column A, the only difference is the amount of non-toxic and toxic examples present within the subgroup.
In column B, the identity subgroup has fewer non-toxic examples and more toxic examples than in column A, and in column C it is the opposite. 
While the extent of unintended bias is clearly the same in all three of these cases, Pinned AUC reports reduced bias in the second case (i.e. higher Pinned AUC value), and increased bias in the final case.
This undesirable distortion of Pinned AUC occurs with any dataset where the balance between classes varies across identity subgroups, making Pinned AUC only applicable on perfectly balanced datasets.

\begin{table*}
\centering
\begin{tabular}{ | >{\centering\arraybackslash} m{2cm} | >{\centering\arraybackslash} m{3.5cm} | >{\centering\arraybackslash} m{3.5cm} | >{\centering\arraybackslash} m{3.5cm} | }
\hline
& A & B & C\\
& Same Toxicity Distribution & More Toxic Examples in Subgroup & More Non-Toxic Examples in Subgroup \\
\hline
Pinned AUC & 0.94 & 0.97 & 0.90 \\
\makecell{Background \\Score \\ Distributions \\ \\ \\ \\ \\ \\ Identity \\ Subgroup \\ Score \\ Distributions} & \includegraphics[scale=0.5]{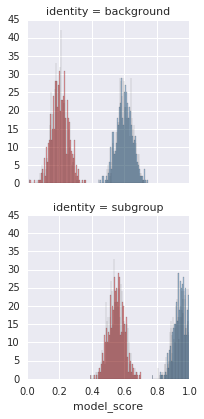} & \includegraphics[scale=0.5]{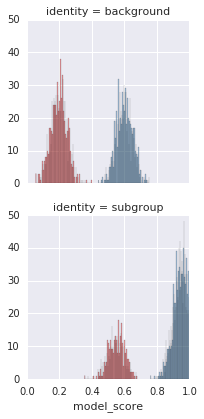} & \includegraphics[scale=0.5]{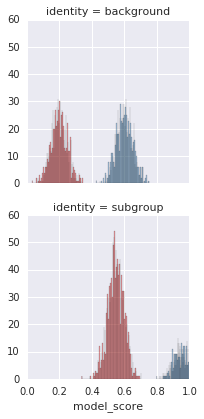} \\
\hline
\end{tabular}
\caption{Examples of score distributions demonstrating bias for a subgroup, and the Pinned AUC metric being affected by background class distribution.}
\label{table:dist}
\end{table*}

The synthetic dataset introduced alongside Pinned AUC in \cite{aies_2018} does have equal distributions between in-class and out-of-class examples, so Pinned AUC provides good insight into unintended bias in that case.
However, synthetic data only provides a very limited view of model performance and is challenging to make realistic synthetic data with broad coverage. 
On the other hand, real data rarely has equal balance between classes across different demographic subgroups.
Therefore, Pinned AUC's requirement of perfectly balanced test sets puts a drastic limitation on Pinned AUC's value to the field as a measure of unintended bias, making it mostly impossible to apply to realistic datasets.

\subsection{Decomposing Pinned AUC}

To understand why Pinned AUC can be distorted by uneven distributions between in class and out of class examples, we break Pinned AUC down into its component parts using the Mann-Whitney $U$ \cite{auc} test statistic. The receiver operating characteristic area under curve ROC-AUC
(which we'll refer to simply as AUC for brevity) can be expressed using the Mann-Whitney U test statistic as:

\begin{equation}
\mbox{AUC}(D) = \frac{\mbox{MWU}(D^{-}, D^{+})}{|D^{-}||D^{+}|}
\end{equation}

As stated in \cite{aies_2018}, Pinned AUC is AUC on a merged dataset, $pD_g$, combining data from the subgroup, $D_g$, with background data, $D$. Using the relationship between AUC and Mann-Whitney $U$, Pinned AUC can be broken down into a combination of four Mann-Whitney $U$-based measures. For simplicity of notation, let $N = |(D_g + D)^-| |(D_g + D)^+|$ or the total number of in-class/out-of-class pairs within the pinned dataset. Then:

\begin{equation}
\begin{split}
\mbox{AUC}(pD_g) & = \mbox{AUC}(D_g + D) \\
& = \frac{\mbox{MWU}((D_g + D)^-), (D_g + D)^+)}{N} \\
& = \frac{\mbox{MWU}(D^-, D^+)}{N} + 
    \frac{\mbox{MWU}(D_g^-, D_g^+)}{N} \\
& + \frac{\mbox{MWU}(D^-, D_g^+)}{N} + 
    \frac{\mbox{MWU}(D_g^-, D^+)} {N} \\
\end{split}
\label{pinned_auc_mwu}
\end{equation}

Using the relationship between the Mann-Whitney $U$ and AUC, Equation~\ref{pinned_auc_mwu} can be rewritten in Equation~\ref{pinned_auc_weighted_avg} as a weighted average of AUCs, where each AUC is weighted by the proportion of example pairs that fall within that group.

\begin{equation}
\begin{split}
\mbox{AUC}(pD_g) & = \frac{|D^-| |D^+|}{N} \mbox{AUC}(D^- + D^+) \\
&+ \frac{|D_g^-| |D_g^+|}{N} \mbox{AUC}(D_g^- + D_g^+)  \\
&+ \frac{|D^-| |D_g^+|}{N} \mbox{AUC}(D^- + D_g^+) \\
&+ \frac{|D_g^-| |D^+|}{N} \mbox{AUC}(D_g^- + D^+) \\
\end{split}
\label{pinned_auc_weighted_avg}
\end{equation}

Viewing Pinned AUC in this decomposition, it becomes clear that if the number of examples in each of the four AUC calculations varies, the overall Pinned AUC value will vary as well.
A main benefit of ROC-AUC is that it is resilient to class imbalances within the test set, but the ``pinning'' technique used to create Pinned AUC does not maintain this benefit.
\cite{aies_2018} recommends sampling such that the background and identity group each make up 50\% of the pinned set, which ensures that the first two terms in Equation~\ref{pinned_auc_weighted_avg} will contain the same number of examples.
\cite{aies_2018} also assumes that the base dataset has identical proportions of in-class and out-of-class examples for all identity subgroups.
When this assumption does not hold, the final two terms in Equation~\ref{pinned_auc_weighted_avg} will be calculated from a different number of example than the other terms, resulting a distorted measurement of unintended bias. 

%% file: 70_synthetic_data_example.tex
\section{Experimental Results}

We further demonstrate the shortcomings of Pinned AUC using the publicly available toxicity classifiers provided by the Perspective API (\cite{perspective_api}) and the synthetically generated test set introduced alongside Pinned AUC in \cite{aies_2018}.

\subsection{Models}
We compare two versions of Perspective API's toxicity classifier, the initial TOXICITY@1 and the latest TOXICITY@6. 
We've previously shown that TOXICITY@1 has been shown to have significant unintended bias around identity words like "gay" and "transgender".
TOXICITY@6 has undergone bias mitigation techniques presented in \cite{aies_2018} and \cite{fp_blog}, and therefore we expect to see reduced unintended bias between these two models as measured by Pinned AUC.

\subsection{Synthetic Test Set} 

\cite{aies_2018} used a template-based synthetic test set with Pinned AUC to demonstrate the reduction in unintended bias between two models similar to TOXICITY@1 and TOXICITY@6. 
We repeat that experiment here, artificially creating a scenario to highlight the weakness in Pinned AUC, causing it to obscure the bias we know to be present. 

As described above, Pinned AUC results are distorted when the dataset does not have equal distribution between in-class and out-of-class data across all evaluated subgroups. 
The synthetic test set \emph{does} have equal distributions, so to demonstrate that distortion we create a second version of the test set where one identity term, ``gay'', has been explicitly skewed to have a lower percentage of non-toxic examples. 
We choose ``gay'' because this term was one of the strongest motivators for the bias mitigation presented in \cite{aies_2018}, and because a low percentage of non-toxic examples containing the word ``gay'' is common in real data, also shown in \cite{aies_2018}.

The original dataset contains 77k examples generated from templates using 50 identity terms, 50\% toxic and 50\% non-toxic across all terms.
The term ``gay'' appears in 1,514 comments. 
To create the skewed dataset, we randomly remove 50\% of the non-toxic examples that include the term "gay", resulting in 1,136 examples including that term, 66\% of which are toxic (with no change to the rest of the dataset).
To mitigate the impact of random noise from the sampling, results for the skewed dataset are averaged over 100 trial runs, where each trial run removes a new random sample of 50\% of non-toxic examples from the identity term ``gay''.

\subsection{Synthetic Test Set Results}

In Table~\ref{tab:synthetic_data_results}, we show Pinned AUC for the 14 identity terms that had original Pinned AUC differences of greater than 0.2 between the two models. 
We compare results between the original evenly balanced test set, and the artificially skewed version of the test set to demonstrate how Pinned AUC fails when data is not perfectly balanced.

The two models, TOXICITY@1 and TOXICITY@6, are the same in the original and skewed scenarios, and therefore the true extent of unintended bias is the same in both scenarios too, the only difference is a removal of a small amount of test data.
An effective metric should report very similar results in these two scenarios, modulo noise due to random sampling to skew the set.
However, the Pinned AUC value for TOXICITY@1 on the skewed identity term, ``gay'', changes markedly from 0.87 to 0.91, seemingly erasing some bias simply by upsetting the balance between toxic and non-toxic data. 
This effect is present within TOXICITY@6 as well, but is less strong.

\begin{table}
\centering
\begin{tabular}{| l | c c || c c |}
\hline
& \multicolumn{2}{|c||}{\textbf{Original}} & \multicolumn{2}{c|}{\textbf{Skewed}} \\
& \multicolumn{2}{|c||}{\textbf{Test Set}} & \multicolumn{2}{c|}{\textbf{Test Set*}} \\
\hline
 & \multicolumn{2}{c||}{Pinned AUC} & \multicolumn{2}{c|}{Pinned AUC} \\
subgroup       & @1 & @6    & @1 & @6   \\
\hline
gay*            & {\cellcolor{lightgray}0.87} & {\cellcolor{lightgray}\textbf{0.96}} & {\cellcolor{lightgray}0.91} & {\cellcolor{lightgray}\textbf{0.97}} \\
homosexual     & 0.85 & \textbf{0.98} & 0.85 & \textbf{0.98} \\
lesbian        & 0.91 & \textbf{0.99} & 0.91 & \textbf{0.99} \\
transgender    & 0.97 & \textbf{1.00} & 0.97 & \textbf{1.00} \\
heterosexual   & 0.98 & \textbf{1.00} & 0.98 & \textbf{1.00} \\
middle eastern & 0.98 & \textbf{1.00} & 0.98 & \textbf{1.00} \\
canadian       & 0.98 & \textbf{1.00} & 0.98 & \textbf{1.00} \\
mexican        & 0.98 & \textbf{1.00} & 0.99 & \textbf{1.00} \\
american       & 0.98 & \textbf{1.00} & 0.99 & \textbf{1.00} \\
elderly        & 0.98 & \textbf{0.99} & 0.99 & \textbf{1.00} \\
lgbt           & 0.98 & \textbf{1.00} & 0.99 & \textbf{1.00} \\
lgbtq          & 0.98 & \textbf{1.00} & 0.98 & \textbf{1.00} \\
younger        & 0.98 & \textbf{1.00} & 0.98 & \textbf{1.00} \\
white          & 0.98 & \textbf{1.00} & 0.99 & \textbf{1.00} \\
\hline
\multicolumn{5}{|p{6cm}|}{*Identity term ``gay'' is the single term that has been skewed
in the skewed set. Results for the skewed set are averaged over 100 trials. } \\
\hline
\end{tabular}
\caption{Comparison between TOXICITY@1 and TOXICITY@6 using the original, evenly distributed, and skewed versions of the synthetic test set. Results in \textbf{bold} show reduction in unintended bias between TOXICITY@1 and TOXICITY@6. Grey background cells highlight the failings of Pinned AUC on skewed data.}
\label{tab:synthetic_data_results}
\end{table}

%% file: 90_conclusion.tex
\section{Conclusion}

This report highlights that Pinned AUC is ineffective at measuring bias in scenarios where the test data does not have identical class distributions between identities.
Moreover, class imbalances are common in real data, and negatively impact Pinned AUC's ability to measure unintended bias. 
However, Pinned AUC can be decomposed into the threshold agnostic metrics presented in \cite{fates-2019} and these variations are robust to changes in data distributions. 
Moreover, these metrics also provide a more detailed view of unintended bias than Pinned AUC, and thus can be used instead to provide a more general framework for measuring unintended bias.

%% file: main.bbl
\begin{thebibliography}{1}

\bibitem{fates-2019}
D.~Borkan, L.~Dixon, J.~Sorensen, N.~Thain, and L.~Vasserman.
\newblock Nuanced metrics for measuring unintended bias with real data for text
  classification.
\newblock In {\em Proceedings of the 1st Workshop on Fairness, Accountability,
  Transparency, Ethics, and Society on the Web}, 2019.

\bibitem{aies_2018}
L.~Dixon, J.~Li, J.~Sorensen, N.~Thain, and L.~Vasserman.
\newblock Measuring and mitigating unintended bias in text classification.
\newblock In {\em Proceedings of AAAI/ACM Conference on Artificial
  Intelligence, Ethics, and Society}, 2018.

\bibitem{perspective_api}
{Jigsaw}.
\newblock Perspective api, 2017.

\bibitem{fp_blog}
{Lucy Vasserman, John Li, CJ Adams, Lucas Dixon}.
\newblock Unintended bias and names of frequently targeted groups, 2018.

\bibitem{auc}
S.~J. Mason and N.~E. Graham.
\newblock Areas beneath the relative operating characteristics (roc) and
  relative operating levels (rol) curves: Statistical significance and
  interpretation.
\newblock {\em Quarterly Journal of the Royal Meteorological Society},
  128(584):2145--2166.

\end{thebibliography}
